\documentclass[a4paper]{article}

\usepackage[english]{babel}
\usepackage[utf8]{inputenc}
\usepackage{amsmath}
\usepackage{graphicx}
\usepackage[colorinlistoftodos]{todonotes}
\usepackage{amssymb,amsthm}
\usepackage{hyperref}

\title{On Generalization and Regularization \\
in Deep Learning\\
\begin{large}
An Introduction for Data Scientists
\end{large}
}
\author{Pirmin Lemberger\\
{\small \texttt{pirmin.lemberger@weave.eu}}
\\
\\
   {\small \textit{Weave Business Technology}}\\
   {\small 37 rue du Rocher, 75008 Paris} \\
   {\small \texttt{weave.eu}}\\ \\
   }
\date{\today}

\begin{document}
\maketitle

\newcommand{\xfeatures}{x_1 \dots, x_p}
\newcommand{\err}{\mathrm{err}}
\newcommand{\errh}{\widehat{\mathrm{err}}}
\newcommand{\Ave}{\mathsf{Ave}}
\newcommand{\sample}{(\bx_1,y_1),\dots,(\bx_n,y_n)}
\newcommand{\wcomp}{w_1, \dots, w_k}
\newcommand{\hf}{\hat{f}}
\newcommand{\E}{\mathsf{E}}
\newcommand{\Rad}{\mathsf{Rad}}
\newcommand{\F}{\mathcal{F}}
\newcommand{\Hs}{\mathcal{H}}
\newcommand{\bx}{\mathbf{x}}
\newcommand{\bw}{\mathbf{w}}
\newcommand{\dd}{\mathrm{d}}
\newcommand{\ES}{\widehat{\E}_S}


\begin{abstract}
Why do large neural network generalize so well on complex tasks such as image classification or speech recognition? What exactly is the role regularization for them? These are arguably among the most important open questions in machine learning today. In a recent and thought provoking paper \cite{understanding_DL} several authors performed a number of numerical experiments that hint at the need for novel theoretical concepts to account for this phenomenon. The paper stirred quit a lot of excitement among the machine learning community but at the same time it created some confusion as discussions in \cite{open_review} testifies. The aim of this pedagogical paper is to make this debate accessible to a wider audience of data scientists without advanced theoretical knowledge in statistical learning. The focus here is on explicit mathematical definitions and on a discussion of relevant concepts, not on proofs for which we provide references.
\end{abstract}

\section{Motivation}
The heart and essence of machine learning is the idea of \emph{generalization}. Put in intuitive terms this is the ability for an algorithm to be trained on samples $S=\{(\bx_1, y_1), \dots, (\bx_n, y_n)\} $ of $n$ examples of an unknown relation between some features $\bx=(\xfeatures)$ and a target response $y$ and to provide accurate predictions on new, unseen data. To achieve this goal an algorithm tries to pick up an estimator $\hf$ among a class of function $\F$, called the \emph{hypothesis class} hereafter, which makes small prediction errors $\vert y_i-\hf(\bx_i) \vert$  when evaluated on the train set $S$. Of course such an endeavor makes no sense unless we can ascertain that the train set error is close in some sense to the true error we would make on the whole population. Practically this discrepancy is estimated using validation and test sets, a procedure familiar to any data scientist.
\\
\par
Again, on intuitive grounds we expect that in order to make good predictions we need to select a hypothesis class $\F$ that is appropriate for the problem at hand. More precisely we should use some prior knowledge about the nature of the link between between the features $\bx$ and the target $y$ to choose which functions the class $\F$ should possess. For instance if, for any reason, we know that with high probability the relation between $\bx$ and $y$ is approximately linear we better choose $\F$ to contain only such functions $f_{\bw}(\bx)=\bw\cdot\bx$. In the most general setting this relationship is encoded in a complicated and unknown probability distribution $P$ on labeled observations $(\bx, y)$. In many cases all we know is that the relation between $\bx$ and $y$ has some smoothness properties.
\\
\par
The set of techniques that data scientists use to adapt the hypothesis class $\F$ to a specific problem is know as \emph{regularization}. Some of these are \emph{explicit} in the sense that they constrain estimators $f$ in some way as we shall describe in section \ref{bv_tradeoff}. Some are \emph{implicit} meaning that it is the dynamics of the algorithm which walks its way through the set $\F$ in search for a good $f$ (typically using stochastic gradient descent) that provides the regularization. Some of these regularization techniques actually pertain more to art than to mathematics as they rely more on experience and intuition than on theorems.
\begin{figure}[h]
\centering
\includegraphics[scale=0.5]{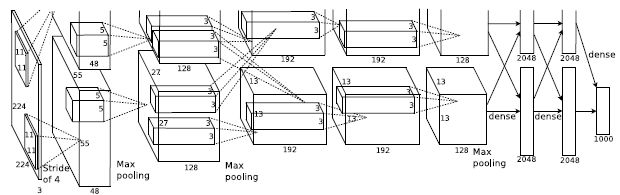} 
\caption{{\small The architecture of \textsf{AlexNet} which is one of the networks used by the authors in \cite{understanding_DL}}}
\label{mlp} 
\end{figure}
\par
\emph{Deep Learning} is a a very popular class of machine learning models, roughly inspired by biology, that are particularly well suited for tackling complex, AI-like tasks such as image classification, NLP or automatic translation. Roughly speaking these models are defined by stacking layers that, each, combine linear combinations of the input with non-linear activation functions (and perhaps some regularization). We won't enter into defining them in detail here as many excellent textbooks \cite{Bishop, Bengio} will do the job. Figure \ref{mlp} shows the architecture of \textsf{AlexNet} a deep network used in the experiment \cite{understanding_DL}. For our purpose, which is a discussion of the issue of generalization and regularization, suffice it to say here that these Deep Learning problems share the following facts:
\begin{itemize}
\item The number $n$ of samples available for training these networks is typically much smaller than the number $k$ of parameters $\bw=(\wcomp)$ that define the functions $f_{\bw}\in\F$ \footnote{The number of parameters $k$ of a Deep Learning network such as \textsf{AlexNet} can be over a hundred of millions while being trained on ``only'' a few millions of images in \textsf{image-net}.}.
\item The probability distribution $P(\bx,y)$ is impossible to describe in any sensible way in practice. For concreteness, think of $\bx$ as the pixels of and image an of $y$ as the name of an animal represented in the picture. Popular wisdom tells us  nevertheless that conditional distributions $P(\bx | y)$ are located in the vicinity of some sort of ``manifold" $M_y\subset\mathbb{R}^p$.
\item The hypothesis class $\F$ is defined implicitly by some neural \emph{network architecture} together with one or more regularization procedures. In section \ref{Experiments_section} we shall briefly describe the more common such procedures. 
\end{itemize}
As a matter of fact, from a practitioner's point of view, Deep Learning with its many tips and tricks \cite{tips_and_tricks} generalizes almost unreasonably well. The question thus is: why? There are two kind of attempts at explaining this mystery.
\begin{enumerate}
\item On the one hand there are heuristic explanations, which rely mostly on variants of physicist's multi-scale analysis \cite{cheap_DL}. Proponents of these explanations argue that the efficiency of deep learning models $\F$ has its roots in the physical generating process $P$ of features $\bx$ (such as image pixels) that describe objects found in nature. They claim this process is generally a hierarchy of simpler processes and that the multi-layered is well adapted to learn such signals. We won't touch upon this further here.
\item On the other there are classical results from Empirical Risk Minimization theory \cite{LNinML} which substantiates the idea that the class $\F$ of hypothesis functions should not be too flexible in order for a model to generalize properly. This is our subject.
\end{enumerate}
Numerical experiments in \cite{understanding_DL} are thought provoking because they challenge the second class of explanations which all interpret regularization as a way of limiting some specific measures of class complexity. In the sequel we shall use \emph{hypothesis class} and \emph{model} interchangeably.  
\\
\\
The main results from \emph{Empirical Risk Minimization} (ERM) theory which form the necessary background for interpreting the experimental results in \cite{understanding_DL} are presented in section \ref{ERM}. However, to put things in perspective we first review the concepts of generalization and regularization in the context of the bias-variance trade-off which is more familiar for data scientists. Section \ref{bv_tradeoff} and \ref{ERM} are actually logically independent but the underlying theme of the \emph{rigidity} of a class $\F$ of functions unifies them. In the context of bias-variance trade-off the rigidity of a model is optimized to minimize the so called \emph{expected loss}, whereas in the context of ERM the rigidity of a model allows to bound population errors with sample errors. Finally, section \ref{Experiments_section} explains in what sense the experiments in \cite{understanding_DL} are challenging ERM.
\section{The Bias-Variance Trade-off}   
Readers in a hurry to learn the basics of ERM can skip this section except for definitions (\ref{pop_err_BV}) and (\ref{err_S}).
\label{bv_tradeoff}
\subsection{The Expected Loss}
In the most general setting the relationship between the features $\bx$ and the target $y$ is given by an unknown probability distribution $P$ from which labeled observations $(\bx,y)$ are drawn. Assume we use a function $f\in\F$ from our hypothesis class to make predictions and that we pick a loss function $\ell(y, f(\bx))\equiv\ell_f(\bx,y)$ to measure the error between the true value $y$ and our prediction $f(\bx)$. The \emph{population error} $\err_P[f]$, also called the \emph{expected loss} (EL), can then be defined as
\begin{equation}
\label{pop_err_BV} 
\begin{split}
	\err_P[f] & \equiv \int  \ell(y, f(\bx)) P(\bx, y) \,\dd \bx \, \dd y  \\
                  & \equiv  \E_P[\ell_f].
\end{split} 
\end{equation}
This is the quantity that we would ideally like to minimize as a functional over $\F$. The special case of a \emph{square loss function} $\ell(y,f(\bx))=[y-f(\bx)]^2$ is interesting, not because it is particularly useful in practice, but because it leads to explicit expressions of various quantities. In this special case, the optimal $h$ which minimizes (\ref{pop_err_BV}) is given by the conditional mean
\begin{equation}
	h(\bx)=\int y P(y \, | \bx) \dd y  \equiv\E_P[y|\bx].
\end{equation}
As $P$ will forever remain unknown to us we must contend ourselves with estimates $\hat{h}_S$ of the optimal $h$ based on finite samples $S=\{\sample\}$. An estimate $\hat{h}_S$ can be defined as a minimizer over $f\in \F$ of the sample version of $\err_P[f]$ 
\begin{equation}
\label{err_S}
\begin{split}
	\errh_S[f] &\equiv \frac{1}{n}\sum_{i=1}^n \ell(y_i, f(\bx_i))  \\
                  & \equiv  \ES[\ell_f].
\end{split} 
\end{equation}
In a frequentist perspective, the accuracy of these empirical minimizers $\hat{h}_S\in\F$ can then be defined by taking an average of the EL over train sets $S$. We call it the \emph{Average Expected Loss} (AEL) even if this naming doesn't particularly shine with elegance. We denote the averaging operation over training samples $S$ by $\Ave$ to distinguish it from the expectation $\E_P$ over labeled observations 
\begin{equation}
\label{AEL_def}
	\mathrm{AEL}\equiv\Ave[\err_P[\hat{h}_S]].
\end{equation}
\subsection{The Bias-Variance Decomposition}
It is instructive to examine the AEL for the special case of a square loss function $\ell(y,f(\bx))=[y-f(\bx)]^2$ because it allows an explicit decomposition of the AEL into three components that are easy to interpret. Indeed, let us denote the \emph{Average Expected Square Loss} by AESL, the best possible estimator by $h(\bx)=\E_P[y|\bx]$ and the minimizer over $\F$ of the empirical error $\widehat{\err}_S[.]$ by $\hat{h}_S$. A straightforward application of the above definitions leads to the following decomposition, see for instance \cite{Bishop} 
\begin{equation}
\label{AESL}
	\mathrm{AESL}=(\mathsf{bias})^2 + \mathsf{variance} + \mathsf{noise},
\end{equation}
where
\begin{equation}
\label{decomp_BV}
\begin{split}
(\mathsf{bias})^2  & = \int \left(\Ave\left[\hat{h}_S(\bx)\right]-h(\bx)\right)^2 P(\bx) \dd\bx, \\
\mathsf{variance}  & = \int \Ave\left[\left(\hat{h}_S(\bx)-\Ave[\hat{h}_S(\bx)]\right)^2\right]  P(\bx) \dd\bx, \\
\mathsf{noise}      & = \int [h(\bx)-y]^2 \,P(\bx,y)\dd\bx\,\dd y.
\end{split} 
\end{equation}
The $(\mathsf{bias})^2$ measures how much our predictions $\hat{h}_S(\bx)$ averaged over all train sets $S$ deviate from the optimal prediction $h(x)$. The $\mathsf{variance}$ measures how much our predictions $\hat{h}_S(\bx)$ fluctuate around their average value when the train set $S$ varies. Finally the $\mathsf{noise}$ measures how much the true values $y$ fluctuate around the best possible prediction $h(\bx)$, this is only term which is independent of our predictions $\hat{h}_S(\bx)$.
\subsection{Regularization as Constraints}
\label{regularization_subsect}
Remember that all prediction functions $\hat{h}_S$ belong to some hypothesis class $\F$ that defines our model. We can control the $\mathsf{variance}$ by putting some constraints on functions which belong to $\F$, thus preventing too large fluctuations of $\hat{h}_S$ when training the model on different samples $S$. Stronger constraint means a more rigid model and thus typically a smaller $\mathsf{variance}$ but a larger $(\mathsf{bias})^2$. Regularization of the model amounts to choosing the amount of \emph{rigidity} that will minimize AEL using an optimal trade-off between the $\mathsf{bias}$ and the $\mathsf{variance}$. For the AESL it is simply their sum which should be minimized as (\ref{AESL}) shows. 
\\
\par
If we consider a parametric model, each function $f_\bw\in\F$ is characterized by a vector $\bw\in \mathbb{R}^p$ of parameters and in particular $\hat{h}=f_{ \widehat{\bw}}$. One way then to make the regularization procedure explicit in this context is to put a constraint on $\bw$ such as for instance $\|\bw\|_2<c$ (\emph{Ridge} regularization) or $\|\bw\|_1<c$ (\emph{Lasso} regularization) where the constant $c$ can be tuned for optimal rigidity. To minimize $\errh_S[f_\bw]$ under such a constraint one proceeds the usual way by defining a Lagrange function and look for its minimum
\begin{equation}
\label{err_reg}
	\errh_S^{\mathrm{reg}} [f_\bw] \equiv \errh_S [f_\bw] + \lambda \|\bw\|_p
\end{equation}
One can show that for each constraint $c$ there is indeed a corresponding $\lambda$ for which $\|\widehat{\bw}\|_p<c$. Summarizing the intuition gained so far and denoting $\F_\lambda$ the class of regularized functions
\begin{center}
larger $\lambda$ $\Leftrightarrow$ smaller $c$ $\Leftrightarrow$  stronger constraints $\Leftrightarrow$ \linebreak ``smaller'' $\F_\lambda$ $\Leftrightarrow$ (smaller $\mathsf{variance}$ and larger $\mathsf{bias}$).
\end{center}
If we want to explicit the dependence of our minimizer on the regularization parameter $\lambda$ we use $\hat{h}_{S,\lambda}$.
\subsection{Regularization in Practice}
\begin{figure}[h]
\centering
\includegraphics[scale=1]{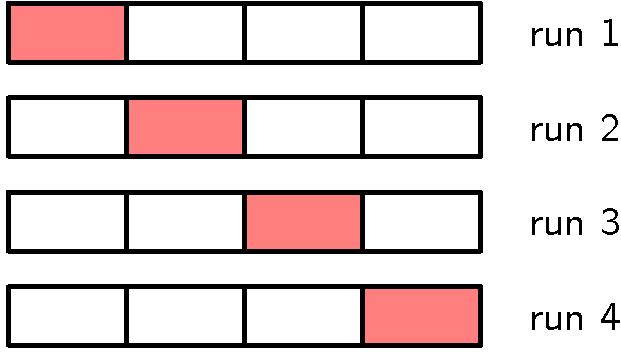} 
\caption{\small{A data set $S$ split into $r=4$ folds. The red rectangles are the test sets $T_j=S\backslash S_j$ associated to the train sets $S_j$.}}
\label{cross_val} 
\end{figure}

Practically the proper amount of regularization $\lambda$ in (\ref{err_reg}) is chosen by using a procedure such as \emph{cross-validation} (CV) well known to data scientist. This procedure can be viewed as giving an estimate $\widehat{\mathrm{AEL}}$ for the true $\mathrm{AEL}$ defined in (\ref{AEL_def}) which involve \emph{two} distinct sampling procedures: 
\begin{enumerate}
\item The first procedure estimates $\Ave$ in (\ref{AEL_def}) by an arithmetic mean over $r$ \emph{train sets} $S_1,...,S_r$ that correspond to a splitting of the original data $S$ into $r$ folds as depicted in figure \ref{cross_val}. 
\item The second procedure estimates the population errors $\err_P[\cdot]$ in (\ref{AEL_def}) by their sample versions $\errh_{T_j}[\cdot]$ on the \emph{test sets }$T_j=S\backslash S_j$.
\end{enumerate}
\begin{figure}
\centering
\includegraphics[scale=1]{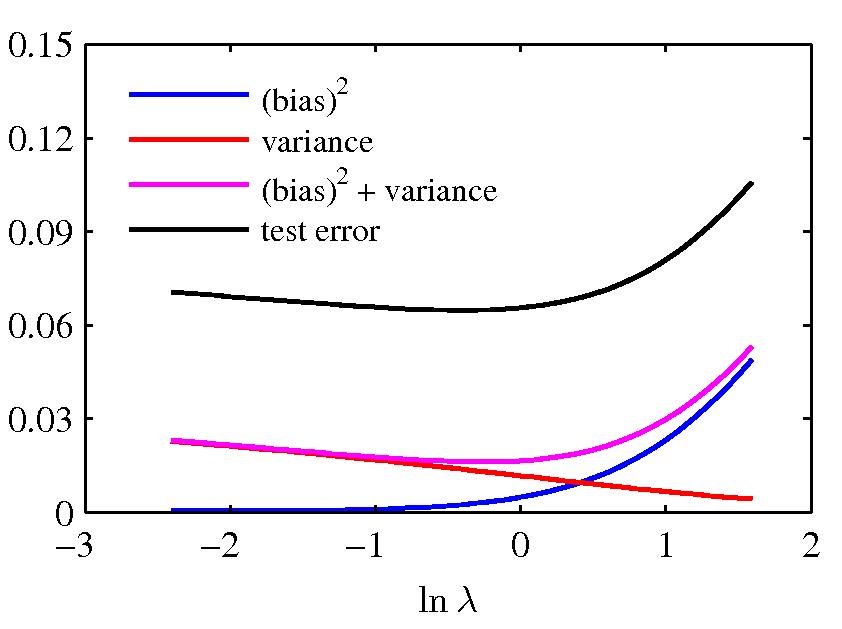} 
\caption{\small{Empirical estimation of the optimal regularization $\lambda$}}
\label{bias_var_tradeoff} 
\end{figure}
We thus define the estimate of $\mathrm{AEL}$ as a function of $\lambda$
\begin{equation}
\label{ESL_est}
	\widehat{\mathrm{AEL}}(\lambda)\equiv\frac{1}{r}\sum_{j=1}^r \errh_{S\backslash S_j}[\hat{h}_{S_j, \lambda}].
\end{equation}
Plotting $\widehat{\mathrm{ESL}}$ as a function of the regularization $\lambda$ allows to find an estimator $\widehat{\lambda}$ of the optimal rigidity of the model. When $r=1$ CV amounts to splitting the data between a train set $S$ and a test set $T$. The \emph{train error} $\errh_S[{\hat{h}_{S,\lambda}}]$ is usually smaller than the \emph{test error} $\errh_T[{\hat{h}_{S,\lambda}}]$ which typically has an U-shaped graph as shown in figure \ref{bias_var_tradeoff}. 
\\
\\
\textbf{Remark 1} : The bias-variance decomposition in (\ref{AESL}) is strictly valid only for a squared loss function $\ell(y, \hat{y})=|y-\hat{y}|^2$. However on experimental grounds the heuristics behind the trade-off, namely the set of equivalences above and the intuition behind figure \ref{bias_var_tradeoff} are expected to remain valid for other loss functions $\ell$ as well.
\\
\\
\textbf{Remark 2} : \emph{Universality theorems} for neural networks tell us that there exists functions $f$, defined by even the simplest neural network (NN) architectures, that will make the sample error $\errh_S[f]$ as small as we want when we minimize \emph{without} regularization constraints ($\lambda=0$). In other words, neural networks can possibly over-fit any sample $S$ provided the network is large enough. These are mere existence theorems however and they tell us nothing neither about how to optimize regularization $\lambda$ to minimize EL nor on our practical ability to find a minimizer in $\F_\lambda$ with these constraints.
\section{Empirical Risk Minimization}
\label{ERM}
The whole analysis in last section rests on our ability to approximate the expected loss $\err_P[f]=\E_P[\ell_f]$ defined in (\ref{pop_err_BV}) with its sample estimate $\errh_S[f]=\ES[\ell_f]$ defined in (\ref{err_S}). But how good exactly is this estimate? This is the question we examine in the current section. It is also the subject matter of \emph{Empirical Risk Minimization} in general. 
\\
\par
We expect the discrepancy between the population expectation $\E_P[h]$ of a function $h$ and its empirical estimate $\ES[h]$ to depend on the class of function $\Hs$ from which we pick $h$. Intuition suggests that if the set $\F$ is too flexible then there are high chances that our empirical estimates could be way off the population expectation. There are several ways to make such hand waving statements rigorous but they all rely on some measure of complexity of a set $\Hs$ of functions. We restrict in this paper to the \emph{Rademacher} complexity for its intuitive character and because its one of the more recent concepts.
\subsection{Rademacher Complexity}
\label{subsection_RC}
To introduce this concept consider a class $\Hs$ of real valued function $h$ defined on an arbitrary space $Z$ and let $S=\{z_1,\cdots,z_n \}$ be a set of examples $z_i$ drawn independently from a distribution $P$ over $Z$. Our aim is to measure how well functions in $\Hs$ are able to match any prescribed binary sequence $\sigma=\{\sigma_1,\dots,\sigma_n\}$ when they are evaluated on samples $S$ drawn from $P$. How well a single function $h$ fits the prescribed sequence $\sigma$ on the sample $S$ can be defined as the correlation $\frac{1}{n}\sum_{i=1}^n \sigma_i h(z_i)$ between $\sigma$ and the values $h(S)\equiv(h(z_1),...,h(z_n))$. It equals $1$ for a perfect fit between $h(S)$ and $\sigma$. How well the class of functions $\Hs$ as a whole can fit a specific sequence $\sigma$ on the sample $S$ can naturally be defined as the highest correlation we can achieve using functions $h\in\Hs$
\begin{equation}
	\label{sup_h}
	\sup_{h\in\Hs} \left(\frac{1}{n}\sum_{i=1}^n \sigma_i h(z_i)\right).
\end{equation}
A measure of how well functions in $\Hs$ can fit \emph{any} sequence $\sigma$ can be defined as the expectation $\E_\sigma$ of (\ref{sup_h}) over sequences $\sigma$ (sampled uniformly). This is by definition the \emph{empirical Rademacher complexity} of the class of functions $\Hs$ on the sample $S$
\begin{equation}
\label{Rad_complexity_emp}
	\widehat{\Rad}_n(\Hs)\equiv\E_\sigma\left[\sup_{h\in\Hs} \left(\frac{1}{n}\sum_{i=1}^n \sigma_i h(z_i)\right)\right].
\end{equation}
Finally, the \emph{Rademacher complexity} of the class $\Hs$ is defined as the expectation of (\ref{Rad_complexity_emp}) over samples $S$ of size $n$ drawn from $P$
\begin{equation}
\label{Rad_complexity}
	\Rad_n(\Hs)\equiv\E_{P}[\widehat{\Rad}_n(\Hs)].
\end{equation}
If we assume that functions $h\in\Hs$ are binary classifiers, which means $h(z)\in \{-1,+1\}$, we see that (\ref{sup_h}) implies $0<\widehat{\Rad}_n(\Hs)\leq 1$. When $\widehat{\Rad}_n(\Hs)$ is close to its upper bound $1$ the binary classification model defined by $\Hs$ can literally ``store'' any binary assignment $\sigma$ to the examples in $S$.  
\subsection{Bounding Population Expectations}
The Rademacher complexity is the key ingredient for bounding a population expectation $\E_P[h]$ with an empirical average $\ES[h]$. Assume we draw samples $S=\{z_1,\dots,z_n \}$ of size $n$ from a distribution $P$ and that the functions $h$ we are interested in belong to a class $\Hs$. Chose a small positive number $0<\delta<1$. Then the following bound holds with a probability larger than $1-\delta$ for any $h\in\Hs$, see \cite{LNinML}
\begin{equation}
\label{basic_bound_S}
	\E_P[h] \leq \ES[h]  + 2\,\widehat{\Rad}_n(\F) + 3\sqrt{\frac{\ln (2/\delta) }{n}}.
\end{equation}
The second term on the right hand side substantiates our intuition that the discrepancy between the population expectation $\E_P[h]$ and the empirical expectation $\ES[h]$ can grow when the set of functions $\Hs$ becomes more and more flexible. The last term is a price we pay for requiring small chances to be wrong, fortunately it grows only logarithmicamlly as $\delta\rightarrow 0$. 
\subsection{Application to Binary Classification Errors}
Let us apply (\ref{basic_bound_S}) to a \emph{binary classification} problem. In this case the set $Z$ is simply the set $\mathbb{R}^p\times \{-1,+1\}$ of observations $z=(\bx, \sigma)$. As $\err_P[f]=\E_P[\ell_f]$ and $\errh_S[f]=\E_S[\ell_f]$ the functions whose expectations we want to bound are the errors function $\ell_f$ associated with binary classifiers $f:\mathbb{R}^p\rightarrow \{-1,+1\}$. Therefore here $\Hs=\{\ell_f | f\in \F \}$ is the class of loss functions. We select the \emph{miss-classification rate} $\ell_f(\bx, \sigma)=\mathbb{I}_{f(\bx)\neq\sigma}$ as our loss function. Using $\mathbb{I}_{f(\bx)\neq\sigma}=[1-\sigma f(\bx)]/2$ and the (almost) obvious property, see \cite{LNinML} 
	\[\widehat{\Rad}_n(a\F + b)=|a|\,\widehat{\Rad}_n(\F)\] 
we get $\widehat{\Rad}_n(\Hs)=\frac{1}{2}\,\widehat{\Rad}_n(\F)$.  Using (\ref{basic_bound_S}) we can thus express our basic inequality directly in terms of the complexity of the hypothesis class $\F$ 
\begin{equation}
\label{basic_ineq_bc}
	\err_P[f] \leq \errh_S[f] + \widehat{\Rad}_n(\F) + 3\sqrt{\frac{\ln (2/\delta) }{n}}
\end{equation}
with probability $>1-\eta$.
\begin{center}
\fbox{\begin{minipage}{25em}
\emph{Inequality} (\ref{basic_ineq_bc}) \emph{is the basic prerequisite  for interpreting the challenging results in paper} \cite{understanding_DL}
\end{minipage}}
\end{center}
\textbf{Remark 1}: It is important to realize that inequality (\ref{basic_ineq_bc}) holds, of course, uniformy for \emph{any} $f\in\F$. In particular $f$ need not be a minimizer of the expected loss which was our main concern in section \ref{bv_tradeoff}.
\\
\\
\textbf{Remark 2}: Inequality (\ref{basic_ineq_bc}) assumes that $f$ is selected from a fixed class $\F$ whereas in the previous section we empirically estimated an optimal rigidity $\widehat{\lambda}$ using a train set $S$ and test set $T$. Although it is very tempting (and morally right!) to substitute $\F_{\widehat{\lambda}}$ for $\F$ in (\ref{basic_ineq_bc}) strictly speaking this is not correct.
\\
\\
For parametric models the next step would be to bound $\widehat{\Rad}_n(\F)$ as a function of the sample size $n$ and the number $k$ of parameters $(w_1,\dots,w_k)$ defining $f_\bw\in\F$. To our knowledge there are no explicit bounds of this kind. There exists however an interesting bound for binary classification models which exemplifies the asymptotic behavior of $\widehat{\Rad}_n(\F)$ as $n\rightarrow\infty$ for a fixed class $\F$. It involves an alternative notion of complexity namely the \emph{VC dimension} $d$ of $\F$ defined as follows: it is the size of the largest set $S=\{\bx_1,\dots,\bx_d\}$ such that for any binary sequence $\sigma=(\sigma_1,\dots,\sigma_d )$ we can find an $f\in\F$ that takes these values on $S$. The bound reads
\begin{equation}
\label{Rad_VC}
	\widehat{\Rad}_n(\F)\leq\sqrt{\frac{2d\ln n}{n}},
\end{equation}
which vanishes when the sample size $n$ grows larger which should come as no surprise.
\section{Regularization for Deep Neural Networks}
\label{Experiments_section}
Now asymptotic behavior is of interest for mathematicians but data scientist are truly  interested in finite samples! Here we can only speculate about the value of $\widehat{\Rad}_n(\F)$ as a function of the size $n$ of the sample and the number $k$ of parameters. Or we can make experiments. In the context of Deep Learning the case of particular interest is $n\ll k$. Recall the following:
\\
\\
\textbf{Classical interpretation of regularization}: Regularization is an explicit or an implicit constraints on predictors $f$ that shrinks the Rademacher complexity of a neural network model $\F$ towards zero so that inequality (\ref{basic_ineq_bc}) nearly saturates.
\\
\\
The most common such mechanisms in Deep Learning are, see \cite{Bishop, Bengio}
\begin{itemize}
\item $L^2$ \textbf{penalty on weights} was shortly discussed in \ref{regularization_subsect} 
\item \textbf{Dropout} consists in randomly dropping some links of the neural network during training thus preventing the model to adapt to closely to the train data.
\item \textbf{Early stopping} consists in monitoring the prediction error on a validation set while performing gradient descent and stopping it when the validation error start to increase. Conceptually it is close to the $L^p$ penalty \cite{Bishop}.
\item \textbf{Data augmentation} is used in image recognition problems. The original data set is augmented with artificial images obtained by distorting (the size, the orientation, the hue of) the original images. It thus acts to increase $n$ without changing $\F$, thus reducing $\widehat{\Rad}_n(\F)$ as (\ref{Rad_VC}) illustrates.
\item \textbf{Implicit regularization} effectively shrinks the class of $\F$ as a consequence of the dynamics of how the algorithm explores $\F$ when optimizing the network parameters $\bw$. Why this happens is more mysterious than for the previous techniques.
\end{itemize}
The numerical experiments described in \cite{understanding_DL} involve large neural networks with a number of parameters $p>1\,000\,000$ such as \textsf{AlexNet} and deep Multilayer Perceptrons. The data sets $S=\{(\bx_1, y_1),\dots, (\bx_n, y_n)\}$ that where used are image data sets such as \textsf{CIFAR-10} which contains $n=60\,000$ images categorized in $10$ classes. The authors also trained the models on pure noise images.
\\
\\
They then created artificial data sets $S^{\sigma}=\{(\bx_1, \sigma_1,\dots, (\bx_n, \sigma_n)\}$ in which the labels $\sigma_i$ where generated randomly. They noticed the surprising fact that for these large networks, \emph{even without any explicit regularization mechanism} switched on, the train error $\errh_{S^\sigma}[h_{S^\sigma}]$ was close to zero ($<0.18\%$) for \emph{any} binary assignment $\sigma$ and even for images made of random pixels. Assuming this multi-class problem is cast into series of equivalent binary classification problems this strongly suggests:
\\
\\
\textbf{Fact 1}: The complexity $\widehat{\Rad}_n(\F)$ of these large models is very close to $1$. In other words these large networks are able to ``learn'' or ``store'' these artificial data sets almost perfectly. But remember:
\\
\\
\textbf{Fact 2}: Deep Learning models are known to generalize very well in practice. This is the reason why people use them! In other words minimizers of the sample error $\errh_S[.]$ are also verified to usefully approximate minimizers of the true population error $\err_P[.]$.
\\
\\
\textbf{Conclusion}: Considering the basic bound (\ref{basic_ineq_bc}), one way to reconcile fact 1 and fact 2 is simply to give up the classical interpretation of regularization that pretends its role is to shrink the model complexity $\widehat{\Rad}_n(\F)$ towards zero. Another possibility is that we need more refined bounds than (\ref{basic_ineq_bc}). Remember that this inequality assumed strictly nothing about the distribution $P$ from which samples $S$ are drawn. It should be no surprise then that these kind of bounds are way too rough. They really don't take into account \emph{at all} what makes the specificity of deep neural networks! The whole point of using techniques such as deep learning is that the models $\F$ they define are somehow well adapted to the actual samples $S$ that nature produces. What is a stake is better understanding of the subtle interplay between the dynamics of the learning process of a multilayered network, which defines an effective $\F$, and the nature of the physical processes $P$ that generate \emph{physical} samples $S$ on which they are trained. But for the moment deep neural networks keep their mystery. 

\section*{Acknowledgments}
I would like thank my colleagues Olivier Reisse, the founder of Weave Business Technology entity, and Christophe Vallet who is leading the Weave Data entity for their consistent support for all activities at the Weave data lab where I work. I would also like to thank Marc-Antoine Giuliani who is an expert in statistical learning theory and data scientist at the Weave data lab for pointing our the many intricacies of his field to a somewhat carefree physicist like me.

\end{document}